%% file: main.tex
\documentclass[10pt]{article} 

\usepackage[hidelinks]{hyperref}
\usepackage{enumitem}
\usepackage{tabularx}
\usepackage{adjustbox}
\usepackage{comment}
\usepackage{geometry} 
\usepackage{comment}

\usepackage{authblk} 

\usepackage[accepted]{tmlr}

\input{math_commands.tex}

\usepackage{url}

\usepackage{microtype}
\usepackage{graphicx}
\usepackage{booktabs} 

\usepackage{algorithm}
\let\classAND\AND
\let\AND\relax
\usepackage{algorithmic}

\let\AND\classAND
\AtBeginEnvironment{algorithmic}{\let\AND\algoAND}
\usepackage{amsmath}

\usepackage{amssymb}
\usepackage{mathtools}
\usepackage{amsthm}
\usepackage{multirow}
\usepackage{adjustbox}
\usepackage{booktabs} 
\usepackage{siunitx} 
\usepackage[normalem]{ulem}
\usepackage{subfig}
\usepackage{float}

\usepackage{hyperref}    
\hypersetup{
    colorlinks=true,          
    linkcolor=blue,           
    filecolor=blue,        
    urlcolor=blue,            
    citecolor=blue,          
}

\usepackage[capitalize,noabbrev]{cleveref}

\theoremstyle{plain}

\theoremstyle{definition}

\theoremstyle{remark}

\title{\centering \raisebox{-0.25\height}{\includegraphics[height=1.3em]{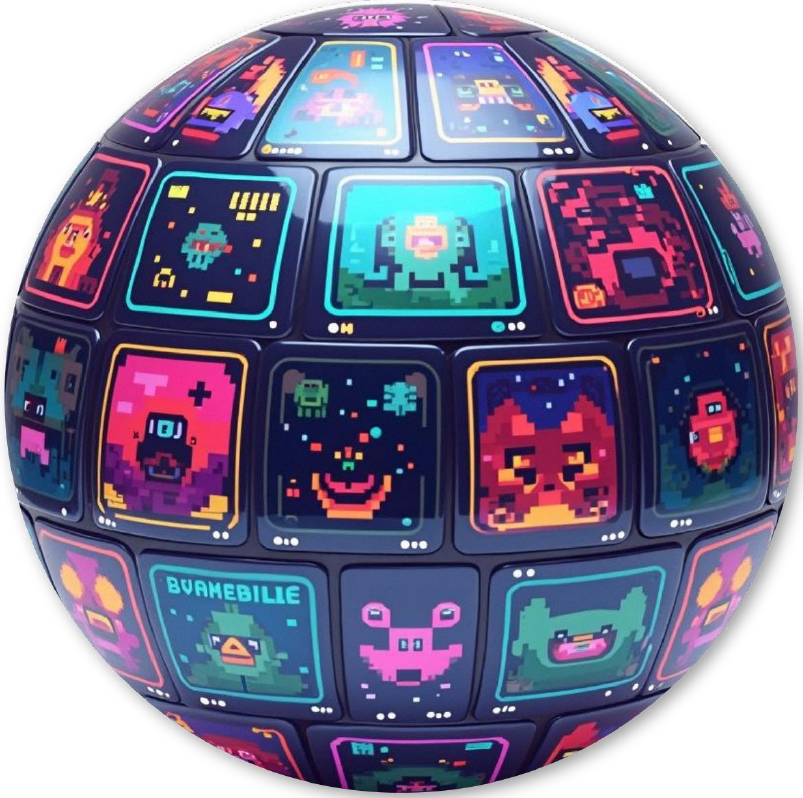}} \textbf{Playable Game Generation}}

\date{}

\author{\normalsize Mingyu Yang, Junyou Li, Zhongbin Fang,  Sheng Chen,  Yangbin Yu, Qiang Fu,  Wei Yang, Deheng Ye}

\affil{Tencent}

\begin{document}

\newgeometry{top=0.1in, bottom=1in, left=1in, right=1in}
\maketitle

\input{section/0_abstract}
\clearpage
\restoregeometry 
\input{section/1_introduction}

\input{section/2_related_work}
\input{section/3_methods}

\input{section/4_experiments}
\input{section/5_conclusion}
\newpage
\bibliography{reference}
\bibliographystyle{tmlr}
\newpage
\input{section/appendix}

\end{document}

%% file: math_commands.tex
\usepackage{amsmath,amsfonts,bm}

\def\eqref#1{equation~\ref{#1}}

\def\1{\bm{1}}

\DeclareMathAlphabet{\mathsfit}{\encodingdefault}{\sfdefault}{m}{sl}
\SetMathAlphabet{\mathsfit}{bold}{\encodingdefault}{\sfdefault}{bx}{n}



%% file: section/0_abstract.tex
\begin{figure}[h]
\centering
\includegraphics[width=0.93\textwidth]{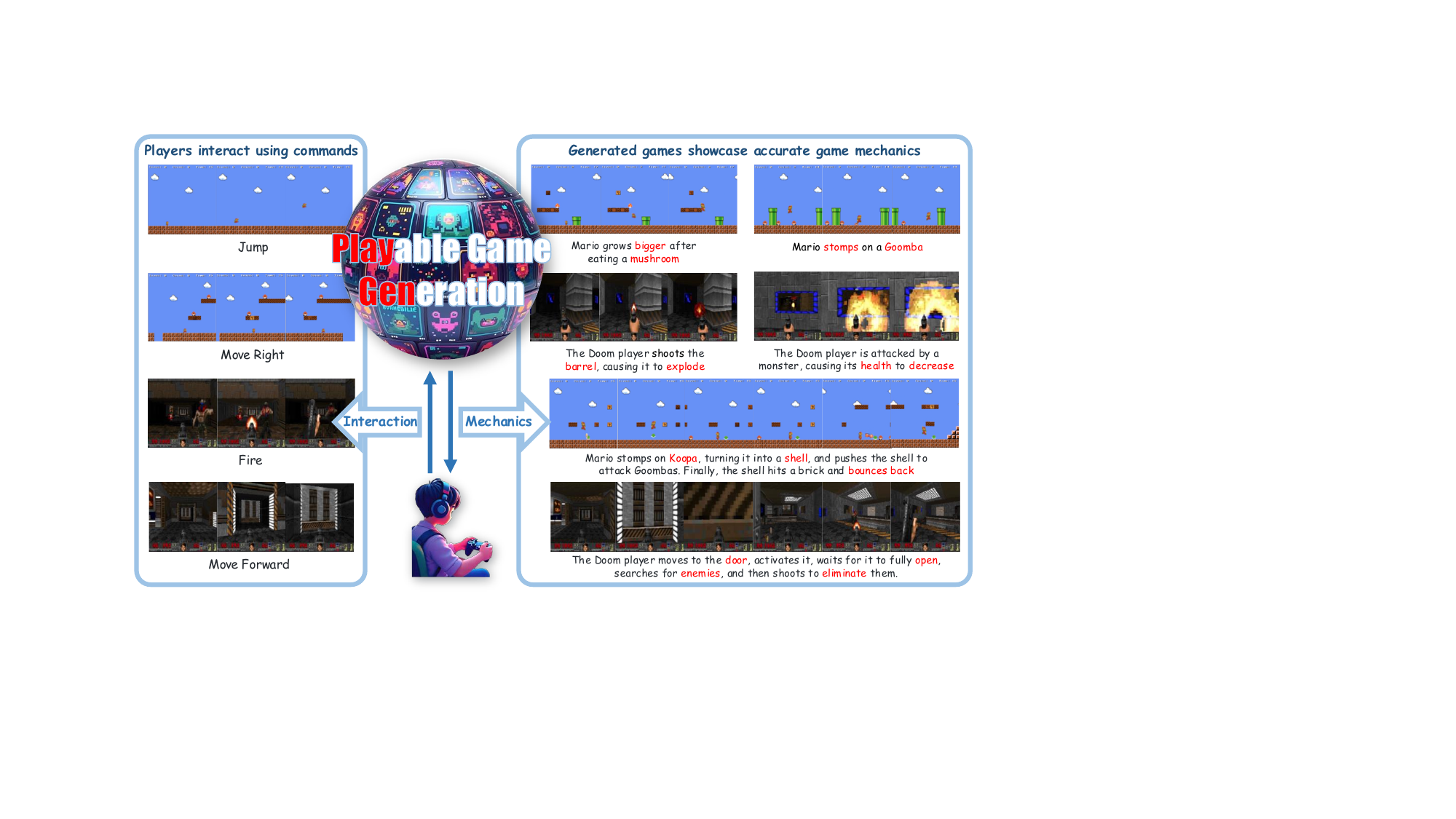}

\caption{
\textbf{Playable Game Generation}: PlayGen can generate playable games that respond to player commands and operate according to game mechanics. Left: we demonstrate how players control in-game characters through commands. Right: we provide cases to show that our method can accurately simulate different game mechanics that involve real-time input actions.}
\vspace{2em}
\label{fig:pr}
\end{figure}

\begin{abstract}

In recent years, Artificial Intelligence Generated Content (AIGC) has advanced from text-to-image generation to text-to-video and multimodal video synthesis. However, generating playable games presents significant challenges due to the stringent requirements for real-time interaction, high visual quality, and accurate simulation of game mechanics. Existing approaches often fall short, either lacking real-time capabilities or failing to accurately simulate interactive mechanics. To tackle the playability issue, we propose a novel method called \emph{PlayGen}, which encompasses game data generation, an autoregressive DiT-based diffusion model, and a comprehensive playability-based evaluation framework. Validated on well-known 2D and 3D games, PlayGen achieves real-time interaction, ensures sufficient visual quality, and provides accurate interactive mechanics simulation. Notably, these results are sustained even after over 1000 frames of gameplay on an NVIDIA RTX 2060 GPU.
Our code is publicly available: \href{https://github.com/GreatX3/Playable-Game-Generation}{here}. 
Our playable demo generated by AI is:  \href{http://124.156.151.207/}{here}.


\end{abstract}


%% file: section/1_introduction.tex
\section{Introduction}

In the recent epoch, Artificial Intelligence Generated Content (AIGC) has undergone a remarkable evolution~\citep{ma,aga}, expanding its capabilities from text-to-text transformations \citep{llm_survey}, through text-to-image synthesis \citep{stable_diffusion,dalle}, to the frontiers of text-to-video \citep{sora} and multimodal video generation \citep{sound_guided,motion_guided,pose_guided}. This progression prompts a fundamental question: What lies beyond these achievements? Games demand not only high generation efficiency and visual quality but, more critically, the accurate simulation of interactive mechanics --- a challenge that surpasses the requirements of previous AIGC outputs. 
The question thus arises: Is game generation a feasible endeavor?

Emerging research has begun to tap into the potential of game generation, yet significant challenges remain. For instance, Genie \citep{genie} endeavors to convert 2D images into interactive games but falls short in supporting real-time interaction, capping interaction duration at a mere 32 frames and lacking meaningful action representation. MarioVGG \citep{mariovgg} employs a pre-trained video generation model, using text as actions within \emph{Super Mario Bros} \citep{mario_dataset} game, yet it also fails to achieve real-time interaction. GameNGen \citep{gameNgen} manages real-time interaction at 20 frames per second (FPS) within the \emph{Doom} \citep{vizdoom} game, but it neglects the simulation of interactive mechanics due to its limited data exploration and memory, leading to significant hallucinations during extended gameplay. 
Collectively, these efforts either lack real-time interaction capabilities or falter in accurately simulating game interactions, culminating in games that are not truly playable.

Playability, as we define it, encompasses three critical attributes: real-time interaction that allows the player's inputs (i.e., actions), sufficient visual quality, and precise simulation of interactive mechanics. Notably, the accurate simulation of interactive mechanics is the cornerstone of playability and is a more pressing and unresolved challenge compared to enhancing visual quality or frame rates. These attributes must be sustained throughout extended gameplay, spanning several minutes.

This realization propels us to devise a game generation methodology that addresses the issue of playability. However, this quest presents a multifaceted challenge: designing methods and model architectures that ensure real-time interaction with arbitrary user actions, while maintaining high visual quality and accurately simulating interactive mechanics that can follow the basic game rules. Moreover, the evaluation of playability remains an uncharted territory in current research.

To surmount these challenges, we propose the following strategies. For generation efficiency and visual quality, we opt for autoregressive diffusion model \citep{diffusion_forcing} over the commonly used but less efficient large language models (LLMs). And we shift the learning objective of the diffusion model from images to VAE-compressed latent vectors and employ DiT \citep{dit} as the network architecture, thereby enhancing generation efficiency and visual quality. 
To accurately simulate interactive mechanics, we tackle the problem primarily from the data perspective, complemented by model architecture enhancements. On the data side, 
we collect a diverse dataset to ensure comprehensive coverage of interaction mechanics. We then balance the collected data to foster unbiased learning and propose a self-supervised long-tailed sample learning method to enhance the simulation of rare interactions. On the model side, we employ an RNN-like model architecture that theoretically possesses infinite memory, ensuring extended gameplay. Lastly, we introduce a playability evaluation method to quantitatively assess interaction efficiency, visual quality, and the fidelity of interactive mechanics simulation.

We validate our approach using the widely recognized 2D game \emph{Super Mario Bros} \citep{java_mario} and 3D game \emph{Doom} \citep{vizdoom}. 
Our findings demonstrate that our method achieves real-time interaction on a consumer-grade NVIDIA RTX 2060 graphics card, ensuring adequate visual quality. Even after over 1000 frames of gameplay, it maintains a precise simulation of interactive mechanics. Our generated games can balance visual quality and frame rate by adjusting parameters, such as denoise sampling timesteps without compromising the accuracy of interactive mechanics. We find that setting the denoise sampling timesteps to 4 nearly doubles the interaction speed, with only a minor decrease in visual quality (1.4\% to 1.8\%) and a negligible reduction in the accuracy of interactive mechanics (0.2\%).

To sum up, our contributions are as follows:
\begin{itemize}
    \item Playability Workflow Development. We develop PlayGen to enable playable game generation with real-time interaction, sufficient visual quality, and accurate simulation of interactive mechanics.
    \item Automated Evaluation Method. We develop an automated evaluation system capable of assessing the generation efficiency, visual quality, and the accuracy of simulating interactive mechanics for game generation models, streamlining the evaluation process.
    \item Benchmarking. Our method has undergone extensive testing on well-established 2D and 3D game benchmarks, demonstrating its effectiveness and reliability. 
\end{itemize}

%% file: section/2_related_work.tex
\section{Related Work}

We first review the most related works which can be categorized into simulating games and converting media into playable games.
Next, we discuss related work on video generation, which employs similar model architectures to PlayGen but with different objectives.


\paragraph{Game Simulation with Neural Networks}
Works in the first category aim to simulate games using neural networks. MarioVGG \citep{mariovgg} generates game video segments of \emph{Super Mario Bros} \citep{mario_dataset} from text-based actions and the first frame of the game. However, it lacks real-time capabilities, taking 4 seconds to generate 6 frames on an NVIDIA RTX 4090. GameNGen \citep{gameNgen} uses a diffusion model to simulate games at 20 FPS on \emph{Doom} \citep{vizdoom} with a single TPU.  Nevertheless, the data collection method of GameNGen fails to ensure coverage and balance in data distribution, resulting in inaccurate simulation of game interactive mechanics.
Oasis \citep{oasis} is a concurrent work of ours and is able to simulate \emph{Minecraft} \citep{minecraft} at 20 FPS on NVIDIA H100. 
In contrast, PlayGen achieves 20 FPS on both \emph{Doom} and \emph{Super Mario Bros} \citep{java_mario} on a less powerful device (NVIDIA RTX 2060) while maintaining the ability to accurately simulate interactive mechanics over extended gameplay. Specifically, after 1000 frames, the accuracy of interactive mechanics decreases by only 0.2\%, suitable for long-term play. 
More importantly, PlayGen shows the complete process of playable game generation, from data generation, model training to model evaluation.

\paragraph{Media to Playable Games Conversion}

The second category, exemplified by Genie \citep{genie}, converts 2D game images and action sequences into game video segments but lacks real-time interaction and is limited to 32 frames, with actions lacking meaningful impact. Other methods, such as \citep{pe2021, pe2022, pe2024}, convert videos into games, requiring a pre-input action sequence and thus lacking real-time interaction. These limitations render them unplayable for extended periods. PlayGen addresses these issues, offering sustained 20 FPS operation.



\paragraph{Playablity Evaluation}
In addition, a significant gap in the most related works is the absence of a playability evaluation method, with most relying on human evaluation. We introduce the first playability-based evaluation method, automating the assessment of game generation models' efficiency, image quality, and interaction mechanics accuracy. 

\paragraph{Video Generation}

These works generate videos through various control signals, such as text \citep{sora}, images \citep{image_guided}, poses \citep{pose_guided}, sounds \citep{sound_guided}, and motion \citep{motion_guided}. Similar to ours, they all use diffusion models as their model architecture to generate continuous frames. However, there are two main distinctions between these works and PlayGen. From a scenario perspective, these works focus on factors such as resolution, length, and alignment with control signals. In contrast, game generation prioritizes real-time performance and the accurate simulation of interactive game mechanics. From a method perspective, video generation methods do not require real-time performance, often taking several minutes or longer to process, and typically employ larger models and datasets for higher resolution and richer content. In contrast, PlayGen ensures real-time performance while utilizing higher-quality data to accurately simulate interactive mechanics.

%% file: section/3_methods.tex
\section{Method}

\begin{figure}[t]
\centering
\includegraphics[width=0.99\textwidth]{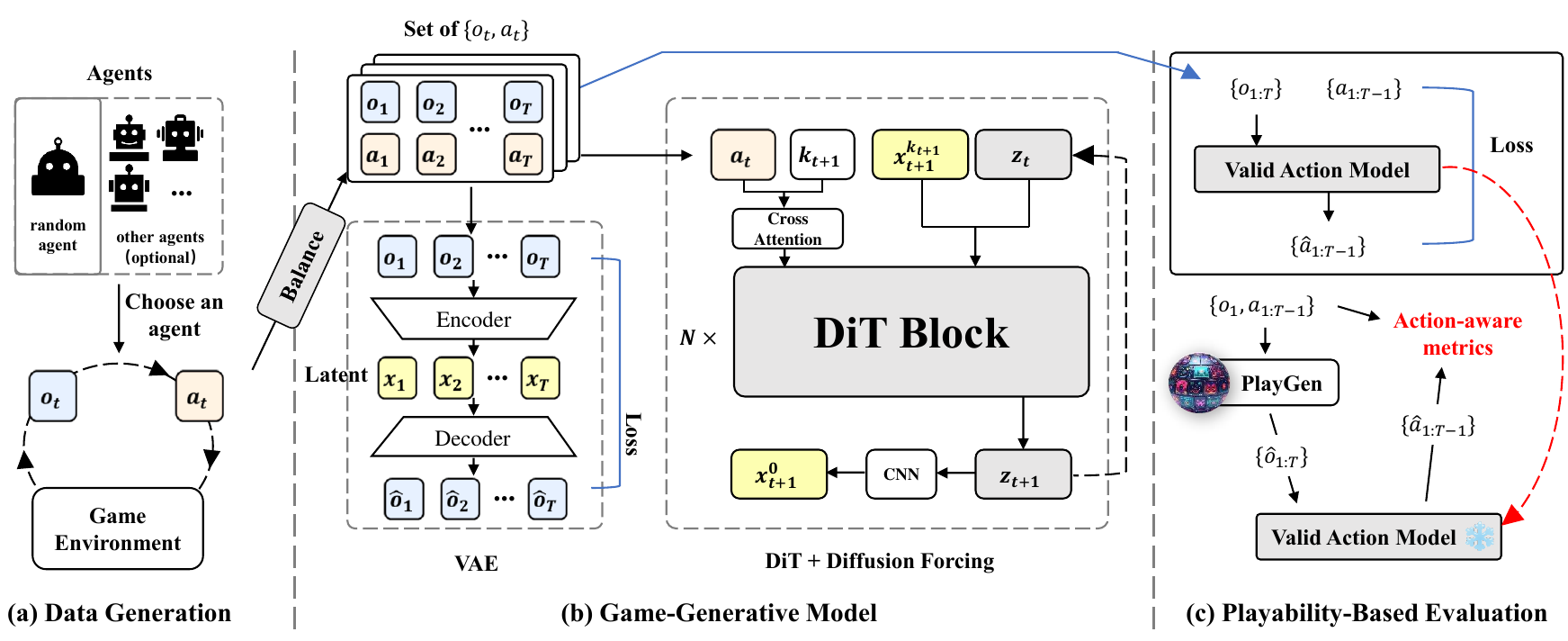}
\caption{
The overall framework of \textbf{PlayGen}. 
\textbf{(a) Data generation.} We generate diverse game data using a hybrid of agents and balance it with cluster-based sampling~(Sec.~\ref{sec:data_processing}).
\textbf{(b) Game-generative model.} Our game-generative model comprises VAE~\citep{vae} and DiT~\citep{dit} with diffusion forcing~\citep{diffusion_forcing}~(Sec.~\ref{sec:model_training}). $a$ and $k$ indicate the current action and the noise level of the next frame respectively. $x$ represents the latent representation encoded by VAE. $z$ is the hidden state of the diffusion model.
\textbf{(c) Playability-based evaluation.} We adopt action-aware metrics calculated via a Valid Action Model~(VAM) to assess the accuracy of interactive mechanics of game-generative models. (Sec.~\ref{sec:model_evaluation})
}
\label{fig:framework}
\end{figure}


\subsection{Overview}

This work leverages a neural network to approximate the game transition model $P(o_{t+1}|o_t, a_t)$, where $o_t$ represents the game observation (i.e., the image rendered by the game) and
$a_t$ represents the action (i.e., the input of user) at timestep $t$. 
With the approximated $P(o_{t+1}|o_t, a_t)$, given any initial game image $o_1$, our network can predict subsequent images $o_{2:n}$ frame by frame as the user inputs actions $a_{1:{n-1}}$ to interacts with the game. This simulates the process of the user playing the game.
To learn $P(o_{t+1}|o_t, a_t)$, we collect game transition data $\{(o_t^i, a_t^i, o_{t+1}^i)\}_{i=1}^m$ to train the neural network that inputs $(o_t^i, a_t^i)$ and outputs $o_{t+1}^i$. In the following,
we detail our method in three main sections:
1) Data Generation: in Sec. \ref{sec:data_processing}, we explain how we generate diverse and balanced transition data, which is essential for training our neural network to simulate a wide range of game scenarios. 2) Model Training and Inference: following data generation, Sec. \ref{sec:model_training} describes the construction of our game-generative model, along with the strategies for training and inference. This section outlines how our model learns to predict game transitions and render images in response to user actions.
3) Model Evaluation: finally, Sec. \ref{sec:model_evaluation} introduces our method for evaluating the playability of the game-generative model. This evaluation is crucial for ensuring that our model meets the standards for real-time interaction, visual quality, and accurate simulation of game mechanics.
Our overall framework, summarizing the flow from data generation to model evaluation, is depicted in Fig. \ref{fig:framework}.

\subsection{Data Generation}

\label{sec:data_processing}

To train a precise transition model of the game, there are two key points for the transition data.
1) Large transition coverage: the observation-action pairs in the training data should cover the whole transition space of the game as much as possible, which can prevent the model from missing some rare but important transitions during training.
2) Balanced transition distribution: the training data is expected to have a balanced distribution within the whole transition space, keeping the model from under-fitting those rare transitions \citep{longtail_survey}. 
To meet these two requirements, we propose a two-stage data generation scheme involving diverse data collection and balanced data sampling.


\paragraph{Diverse Data Collection.}
\label{sec:data_collect}

We employ random agents and various other agents (optional) to interact with the game environment to collect game data. 
The random agent is responsible for exploring specific areas as thoroughly as possible, while other agents aim to progress through the game with the objective of completion. Other agents can be provided by the game engine or trained using reinforcement learning (RL).
Specifically, at each timestep in the game, an agent has a probability $p \in [0, 1]$ of choosing a random action, and a probability $1-p$ of following the decision made by other agents (e.g., expert agents or RL agents). Meanwhile, we repeatedly perform an action one or multiple times if this action is chosen by random agents, which makes the data more human-like. 
Utilizing this method, we can generate transition data that covers most cases of the game. 
During this process, we record the rendered images $o_t$ of the game, the actions $a_t$ taken by the agents, and additional available information $e_t$ within the game (e.g.,  the position of agent). We conclude $T$ timesteps into one sample.
Notably, due to our diverse data collection strategy, we avoid the need to design complex, domain-specific rewards for training RL agents to explore rare observations. And it allows us to employ even a single random agent in highly customizable games, thereby making our approach more general. The pseudo-code of diverse data collection is shown in Algorithm~\ref{alg:data_collection}.

\begin{algorithm}[t!]
\caption{Diverse Data Collection}
\label{alg:data_collection}
\begin{algorithmic}[1]
\STATE \textbf{Input:} Game engine $G$, random agents $A_r$, other optional agents $A_o$ (e.g., RL agents, expert agents), number of episodes $K$, number of timesteps per episode $T$.
\STATE \textbf{Output:} Diverse game transition dataset $\mathcal{D}$.
\FOR{$episode = 1, 2, \cdots, K$}
    \STATE Randomly set the initial state of game $s_0 \sim p(s_0)$.
    \STATE Generate a random number $p \in [0,1]$ as the probability of taking random actions at each timestep of this episode.
    \FOR{$t = 1, 2, \cdots, T$}
        \STATE Choose action $a_t$ with probability $p$ by $A_r$ and with probability $1 - p$ by $A_o$.
        \STATE Record the rendered image $o_t$ of the game, the action $a_t$, and additional available information $e_t$ within the game (e.g., the position of agent).
        \STATE Execute action $a_t$ in $G$.
    \ENDFOR
    \STATE Conclude $T$ timesteps into one sample and store it into $\mathcal{D}$.
\ENDFOR
\end{algorithmic}
\end{algorithm}


\paragraph{Balanced Data Sampling.}
\label{sec:data_balance}

%
%
%
%

The extensive coverage of the data provides examples of nearly all scenarios within the game, but it potentially results in data distribution imbalance. 
%
%
For instance, in \emph{Super Mario Bros}, if Mario encounters a high pipe, it often gets stuck, leading to the accumulation of a lot of data for that scenario, causing data imbalance.
To address this imbalance and achieve a more balanced transition distribution, we propose a cluster-based data sampling method that samples transitions from the collected large-scale dataset in a balanced manner.

Specifically, we first calculate a feature vector based on the additional info $e_t$ for each sample to represent the transition characteristics. 
For example, if $e_t$ saves the position of agent, we can calculate the position distribution over different game areas as the transition characteristics for a sample. 
Our goal is to sample a subset of large-scale data and ensure it has balanced transition characteristics (e.g., balanced position distribution) as a whole. 
To achieve this, we cluster all samples into $k \in \mathbb{N}$ categories based on the feature vectors, and get $k$ cluster centers denoted by $k$ feature vectors $\{\mathbf{c}_1, \mathbf{c}_1, \cdots, \mathbf{c}_k\}$. 
We treat $\mathbf{c}_i$ as the approximate transition characteristics for every sample in the $i^{th}$ cluster. 
Then, we formulate a linear equation as follows:
\begin{equation}
b_1\mathbf{c}_1 + b_2\mathbf{c}_2 + \cdots + b_k\mathbf{c}_k = \mathbf{y},
\end{equation}
where $\mathbf{y}$ is the target transition characteristics, i.e., balanced transition characteristics, and $\{b_1, b_2, \cdots, b_k\}$ is the solution of this linear equation. By leveraging the non-negative least squares method, we obtain an approximate non-negative integer solution $\{b_1, b_2, \cdots, b_k\}, b_i \in \mathbb{N}$. Finally, we sample $b_i$ samples in the $i^{th}$ cluster and get a balanced transition dataset consisting of $\sum_{i=1}^k b_i$ samples. The pseudo-code of balanced data sampling is shown in Algorithm~\ref{alg:data_balance}.

\begin{algorithm}[t!]
\caption{Balanced Data Sampling}
\label{alg:data_balance}
\begin{algorithmic}[1]
\STATE \textbf{Input:} Collected transition dataset $\mathcal{D}$, number of clusters $k \in \mathbb{N}$.
\STATE \textbf{Output:} Balanced transition dataset $\mathcal{D}_{\text{balanced}}$.
\STATE Calculate the transition characteristics (e.g., position distribution) based on $e_t$ as a feature vector for each sample in $\mathcal{D}$.
\STATE Cluster all samples into $k$ clusters based on the feature vectors, and obtain $k$ cluster centers $\{\mathbf{c}_1, \mathbf{c}_2, \cdots, \mathbf{c}_k\}$.
\STATE Formulate a linear equation:
$$
b_1 \mathbf{c}_1 + b_2 \mathbf{c}_2 + \cdots + b_k \mathbf{c}_k = \mathbf{y},
$$
where $\mathbf{y}$ is the target balanced transition characteristics, i.e., balanced transition characteristics (e.g., balanced position distribution).
\STATE Solve the linear equation using the non-negative least squares method to obtain an approximate non-negative integer solution $\{b_1, b_2, \cdots, b_k\}$, $b_i \in \mathbb{N}$.
\FOR{$i = 1, 2, \cdots, k$}
    \STATE Sample $b_i$ samples from the $i^{th}$ cluster.
\ENDFOR
\STATE Obtain $\mathcal{D}_{\text{balanced}}$ consisting of $\sum_{i=1}^k b_i$ samples.
\end{algorithmic}
\end{algorithm}

\subsection{Game-Generative Model}
\label{sec:model_training}

We develop an action-conditioned diffusion model to learn the game transition model $P (o_{t+1}|o_t, a_t)$. Below, we describe our model architecture, and its training and inference strategies.


\paragraph{Model Architecture.}

Our model architecture consists of a Variational Autoencoder \citep{vae} (VAE) and a Latent Diffusion Model (LDM) \citep{stable_diffusion}. 
The VAE consists of an encoder and a decoder, where the encoder encodes the original image $o_t$ into a latent representation $x_t$, and the decoder reconstructs the original image from the latent. 
In many video games, the background part of the game usually remains the same or changes very little within an sample. 
Hence, the latent $x_t$ can be learned with lower dimension and store tighter information than $o_t$, ignoring redundant information in the game background. 
By utilizing the VAE, we shift the objective of learning game transition model from image space to latent space.
We then train an LDM to approximate the game latent transition model $P(x_{t+1}|x_t, a_t)$, which denoises $x_{t+1}$ conditioned on $x_t$ and $a_t$ at timestep $t$.
However, many games cannot be strictly viewed as Markov Decision Processes (MDP), i.e., the next latent $x_{t+1}$ not only depends on current latent $x_t$ and action $a_t$, but also depends on all past latents $x_{1:t-1}$ and actions $a_{1:t-1}$, which we call ``memory''. 
Hence, to accurately denoise the next latent $x_{t+1}$, we need to condition on $x_{1:t}$ and $a_{1:t}$, leading to memory length explosion if timestep $t$ is large.

To maintain long-term memory under limited computing resources, we employ an autoregressive RNN-like model structure \citep{diffusion_forcing} for LDM.
%
Specifically, as shown in Fig.~\ref{fig:framework}(b), we learn a hidden state $z_t$ to capture the influence of current latent $x_t$ and past memory (i.e., $x_{1:t-1}$ and $a_{1:t-1}$). Then, we can denoise $x_{t+1}$ conditioned on $z_t$ and $a_t$. In practice, we concatenate $z_t$ with noisy latent observation $x_{t+1}^{k_{t+1}}$ as the input of LDM, and concatenate embeddings of action $a_t$ and noisy level $k_{t+1}$ as the conditions used for cross attention mechanism.
%
%
The backbone of LDM consists of $N \in \mathbb{N}$ DiT \citep{dit} blocks, which output the next hidden state $z_{t+1} \sim p_{\theta}(z_{t+1}|z_{t}, x_{t+1}^{k_{t+1}}, a_{t}, k_{t+1})$, where $\theta$ are the network parameters.
Next, a convolutional neural network (CNN) $p_\phi(x_{t+1}^0|z_{t+1})$ with parameters $\phi$ is utilized to predict $x_{t+1}=x_{t+1}^0$ from hidden state $z_{t+1}$.
%


\paragraph{Training.}

We train the game-generative model in two stages as follows.
First, we train the VAE using the weighted sum of reconstruction loss, perceptual loss with LPIPS~\citep{lpips} and KL-penalty. 
Then, we freeze the  VAE and train the LDM with the objective of velocity parameterization~\citep{pred_v}. 
Even if the training data are collected in a diverse way and have been balanced, we find that the model still fails in some rare transitions. 
For example, Mario cannot change to fire status after eating a fire flower when we play our trained model that simulates the \emph{Super Mario Bros} game. 
%
This is because data sampling only balance the distribution without increasing the scarce data  (i.e., long-tailed transitions). The model is prone to underfitting these long-tail transitions, resulting in poor performance.
%
%
%
To solve this issue, we present a self-supervised long-tailed transition learning method as follows.


\paragraph{Self-Supervised Long-Tailed Transition Learning.} 

One of the mainstream solutions for long-tailed learning is re-sampling~\citep{longtail_survey}, which increases the sampling probability of tail-class samples during training and alleviates the model’s underfitting on long-tailed samples. 
However, the re-sampling method is unable to be easily extended to long-tailed transition learning since transition data are unlabeled.
To this end, we propose a method for solving the long-tailed problem in self-supervised transition learning. 
Inspired by the Prioritized Experience Replay (PER)~\citep{per} algorithm in RL, we identify transitions with high loss as long-tailed transitions and then increase the training frequency of these transitions. 
%
Specifically, we use a priority queue of size $N_q \in \mathbb{N}$ to maintain the top $N_q$ samples with the highest loss during training. 
At each training step $t$, we sample data from the priority queue with probability $p_t \in [0, 1]$ and from the whole transition dataset with probability $1 - p_t$, 
%
The pseudo-code of self-supervised long-tailed transition learning is shown in Algorithm~\ref{alg:long_tailed_learning}.

\begin{algorithm}[t!]
\caption{Self-Supervised Long-Tailed Transition Learning}
\label{alg:long_tailed_learning}
\begin{algorithmic}[1]
\STATE \textbf{Initialize:} The balanced transition dataset $\mathcal{D}_{\text{balanced}}$, and a priority queue $\mathcal{Q}$ of size $N_q$ as the long-tailed transition dataset.
\FOR{each training step $t$}
    \STATE Compute the moving average loss $\bar{L}$ over the last $M$ training steps.
    \STATE Set $p_t$ that is negatively correlated with $\bar{L}$.
    \STATE Sample a batch of transition data from $\mathcal{Q}$ with probability $p_t$ and from $\mathcal{D}_{\text{balanced}}$ with probability $1 - p_t$. 
    \STATE Update model using the sampled transition data.
    \STATE Update $\mathcal{Q}$ with the sampled transition data based on the loss value, and ensure that $\mathcal{Q}$ maintains the top $N_q$ training samples with the highest loss during training.
\ENDFOR
\end{algorithmic}
\end{algorithm}


\paragraph{Inference.}
After training, we utilize DDIM sampling~\citep{ddim} for inference. With only 4 DDIM sampling timesteps, our game-generative model is able to simulate both video games \emph{Super Mario Bros} and \emph{Doom} at 20 FPS on NVIDIA RTX 2060, which enables real-time interaction for users.

\subsection{Playability-Based Evaluation}
\label{sec:model_evaluation}

%
We define ``playability'' to encompass the following dimensions:
\begin{itemize}
    \item Real-time Interaction. It refers to the ability of the game-generative model to generate and display game frames quickly enough to provide a smooth and uninterrupted gaming experience.
    \item Sufficient visual quality. It means that the generated frames should maintain the same level of detail, color accuracy, and overall visual fidelity as the original game.
    \item Accurate simulation of game interactive mechanics. It indicates that the behavior of characters in the generated frames should accurately reflect the physical rules and interactive mechanics of the game.
\end{itemize}
%
Previous research has predominantly focused on evaluating the first two dimensions, neglecting the assessment of the accuracy of interactive mechanics. In contrast, we choose to address this aspect to comprehensively assess the concept of ``playability''. 
We find that assessing the accuracy of interactive mechanics is equivalent to evaluating the correctness of action execution. For instance, if a jump action is executed in the game, the character should be rendered as jumping in the next frame. If the next frame shows the character moving left instead, the action has not been correctly executed, indicating an error of the interactive mechanics. 
Based on this finding, we design two complementary action-aware metrics to evaluate the accuracy of action execution, detailed as follows.
%



\paragraph{ActAcc Metric.}
We train a Valid Action Model (VAM) to recognize the actions between adjacent frames. By comparing these recognized actions with the actual executed actions, we calculate the accuracy, referred to as the ActAcc metric, to evaluate the interactive mechanics. We define this metric as follows:

\begin{equation}
\text{ActAcc} \coloneqq \frac{1}{L} \sum_{i=1}^{L} \left( a^{\text{pred}}_i == a^{\text{gt}}_i \right),
\end{equation}
where $L$ represents the length of the evaluated sequence, $a^{\text{pred}}_i$ and $a^{\text{gt}}_i$ are the action predicted by the VAM and the ground-truth action of the $i^{th}$ frame, respectively. 

To ensure that the VAM can predict actions without bias, we use the same balanced dataset as the game-generative model for the training set. Its input consists of \(t\) frames before and after the current frame, and it outputs the current action.
%
%
%
The network architecture is inspired by VPT~\citep{vpt}, with the key enhancement being the addition of a spatial-temporal transformer~\citep{st-transformer} at the end of the network. This modification aims to improve the understanding ability of model for spatial-temporal dependencies.



\paragraph{Extend ActAcc with ProbDiff Metric.}

However, we find that the ActAcc metric has inherent limitations because actions may not always affect the subsequent frames. Different actions in the current frame may lead to identical outcomes.
%
%
For example, in the \emph{Super Mario Bros} game, when Mario is in the air, the ``left jump'' action and the ``move left'' action, as well as the ``right jump'' action and the ``move right'' action, may have the same effect. 
In such scenarios, the VAM fails to distinguish between these actions, resulting in inaccurate evaluations when leveraging accuracy (i.e., ActAcc) as a metric.
We find that in this case, since the actions are not distinguishable, the output action probability distribution tends to be uniform. This means that the probability of the predicted action and ground-truth action are very similar.
We leverage this finding to address the shortcomings of the ActAcc metric and propose the ProbDiff metric, defined as follows:
\begin{equation}
\text{ProbDiff} \coloneqq \frac{1}{L} \sum_{i=1}^{L} \left( P(a^{\text{pred}}_i) - P(a^{\text{gt}}_i) \right),
\end{equation}
where $P(a^{\text{pred}}_i)$ and $P(a^{\text{gt}}_i)$ are the output probability of VAM corresponding to the predicted action and the ground-truth action for frame \(i\), respectively.
%
%
When the action is not able to influence the subsequent frames, the probability of these actions is similar, resulting in a minimal increase in ProbDiff metric. This effectively mitigates the limitation of ActAcc metric.
However, a limitation of ProbDiff metric is that if completely unrelated or very low-quality frames are provided, VAM will still output uniform action probability distribution, leading to misleading ProbDiff scores. In contrast, ActAcc will output low scores because VAM fails to recognize the actions, thus remaining unaffected by this case. Therefore, we use both ActAcc and ProbDiff metrics to complement each other in evaluating the accuracy of game interactive mechanics.
%
%

%% file: section/4_experiments.tex
\section{Experiments}


\begin{figure}[t]
\centering
\includegraphics[width=\textwidth]{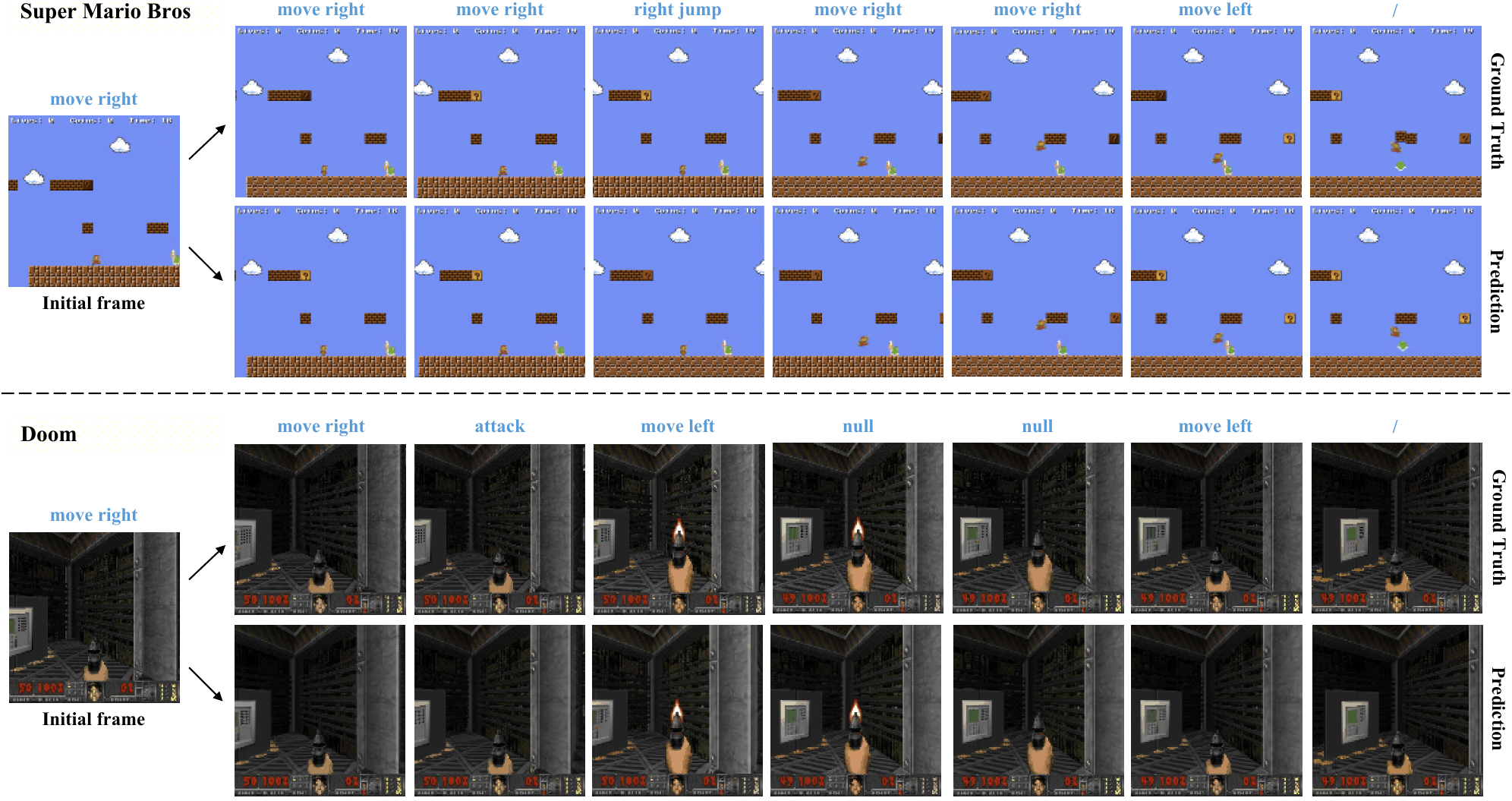}
\caption{
Visualization of the predicted results from PlayGen and the ground truth.
}
\label{fig:prediction_vis}
\end{figure}

In this section, we conduct experiments to verify the effectiveness of PlayGen.
We first show quantitative results that demonstrate PlayGen meets the criteria for playability from three aspects: visual quality, simulation of game interactive mechanics and interaction efficiency. Then, we conduct analysis to show 1) how PlayGen progressively achieves playability with three important components: diverse data collection, balanced data sampling and 
self-supervised long-tailed transition learning, 2) the rationality of action-aware metrics for evaluating the accuracy of game interactive mechanics, and 3) the impact of data balanced sampling.


\paragraph{Experimental Setup.} We validate PlayGen on the widely recognized 2D game \emph{Super Mario Bros} \citep{java_mario} and 3D game \emph{Doom} \citep{vizdoom}.
On the \emph{Super Mario Bros} game environment, we collect 236M frames of transitions based on the Mario-AI-Framework~\citep{java_mario}, and then sample 50M balanced transitions for training. While on the \emph{Doom} game environment, we collect 900M frames of transitions based on the ViZDoom~\citep{vizdoom}, and then sample 200M balanced transitions for training.
We collect $T = 200$ timesteps into one sample.
All frames (during training and testing) are at a resolution of $128 \times 128$.
Only the first frame and subsequent actions from ground-truth trajectories are given as context information for the inference process. 

Also, we train a VAM to calculate ActAcc and ProbDiff metrics. The same balanced dataset is used as the training dataset. The input of VAM is a frame sequence, with the sequence length set to 32. For any given frame in the sequence, the 8 preceding and 8 succeeding frames are used as context frames. Therefore, our action-aware metrics can theoretically only evaluate sequences longer than 17 frames, as VAM requires 16 context frames.

\begin{table}[t]
    \centering
    \renewcommand{\arraystretch}{1.2}
    \setlength{\tabcolsep}{6pt}
    \caption{\textbf{The visual quality and accuracy of game interactive mechanics with different prediction lengths.}  We bold the best results of each metric and underline the instances that need to be focused on. ``-'' means not applicable.}
    \begin{tabular}{llllllll}
        \toprule
        \multirow{2}{*}{Game} & \multirow{2}{*}{Metric} & \multicolumn{6}{c}{Prediction Length} \\
        & & 1 & 16 & 32 & 64 & 128 & \underline{1024} \\
        \midrule
        \multirow{6}{*}{\emph{Super Mario Bros}}
        & LPIPS$\downarrow$ & \textbf{0.022} &  0.043 & 0.062 & 0.083 & 0.122 & 0.222 \\
        & PSNR$\uparrow$ & \textbf{33.81} & 28.06 & 26.18 & 24.56 & 22.59 & 18.19 \\
        & FID$\downarrow$ &  15.24 & 7.09 & \textbf{6.12} & 7.52 & 12.85 & 50.91 \\
        & FVD$\downarrow$ & - & 195.36 & \textbf{105.44} & 109.31 & 123.49 & 173.06 \\
        & ActAcc$\uparrow$ & - & - & \textbf{0.803} & 0.801 & 0.790 & 0.789 \\
        & ProbDiff$\downarrow$ & - & - & \textbf{0.056} & 0.059 & 0.063 & 0.065 \\ 
        \midrule
        \multirow{6}{*}{\emph{Doom}}
        & LPIPS$\downarrow$  & \textbf{0.165} & 0.253 & 0.285 & 0.311 & 0.346 & 0.472 \\
        & PSNR$\uparrow$ & \textbf{23.81} & 21.28 & 20.41 & 20.03 & 19.24 & 17.25 \\
        & FID$\downarrow$ & 76.48 & \textbf{50.37} & 58.75 & 60.86 & 72.44 & 136.40 \\
        & FVD$\downarrow$  & - & \textbf{390.30} & 622.51 & 711.09 & 730.29 & 2176.94 \\
        & ActAcc$\uparrow$  & - & - & \textbf{0.858} & 0.851 & 0.848 & 0.822 \\
        & ProbDiff$\downarrow$  & - & - & \textbf{0.019} & 0.020 & 0.022 & 0.028 \\ 
        \bottomrule
    \end{tabular}
    \label{tab:metrics}
\end{table}

\begin{table}[t]
    \centering
    \renewcommand{\arraystretch}{1.2}
    \setlength{\tabcolsep}{6pt}
    \caption{\textbf{The interaction efficiency correlated to the visual quality and the accuracy of interactive mechanics with different denoise sampling timesteps.} We test on NVIDIA RTX 2060 GPU and bold the best results.}
    \begin{tabular}{c|c c c|c c c}
        \toprule
        \multirow{2}{*}{Metric} & \multicolumn{3}{c|}{\emph{Super Mario Bros}} & \multicolumn{3}{c}{\emph{Doom}} \\
        \cline{2-7}
        & 4 & 8 & 16 & 4 & 8 & 16 \\
        \midrule
        LPIPS$\downarrow$ & 0.064 & 0.062 & \textbf{0.060} & 0.289 & 0.285 & \textbf{0.282} \\
        PSNR$\uparrow$ & 26.53 & 26.18 & \textbf{26.61} & \textbf{20.74} & 20.41 & 20.40 \\
        FID$\downarrow$ & 6.23 & \textbf{6.12} & 6.18 & 78.54 & \textbf{58.75} & 75.37\\
        FVD$\downarrow$ & 107.33 & \textbf{105.44} & 120.30 & 1156.66 & \textbf{622.51} & 1030.68 \\
        ActAcc$\uparrow$ & 0.791 & 0.801 & \textbf{0.808} & 0.834 & \textbf{0.858} & 0.857 \\
        ProbDiff$\downarrow$ & 0.061 & 0.059 & \textbf{0.048} & 0.021 & \textbf{0.019} & 0.019 \\
        \midrule
        FPS $\uparrow$ & \textbf{20} & 10 & 5 & \textbf{20} & 10 & 5 \\
        \bottomrule
    \end{tabular}
    \label{tab:denoise}
\end{table}

\subsection{Quantitative Results}

In this subsection, we present quantitative results regarding the playability of PlayGen. As mentioned, the criteria for playability focus on three aspects: real-time interaction, sufficient visual quality, and accurate simulation of interactive mechanics. We first show the visual quality of generated frames. Then, we evaluate the accuracy of interactive mechanics, which is an important point for playability and has not been studied in previous works. Finally, we test the interaction efficiency and demonstrate that PlayGen can achieve real-time interaction while ensuring visual quality and simulation of interactive mechanics. Moreover, these results are sustained even after over $1000$ frames.

\subsubsection{Visual Quality}
We adopt LPIPS \citep{lpips}, PSNR \citep{psnr}, FID \citep{fid} and FVD \citep{fvd} to measure the visual quality, where FVD is used to evaluate video quality while the other three metrics are used to evaluate image quality. Thus, FVD is not applicable for prediction length of $1$.
We refer to several works~\citep{svd,stable_diffusion} on image and video generation and set thresholds on these metrics, the visual quality is sufficient when $\text{LPIPS} < 0.2, \text{PSNR} > 20, \text{FID} < 85 \text{ and FVD} < 300$. 
We calculate these metrics on $600$ ground-truth trajectories (covering most game scenarios) with different prediction lengths. The longest prediction length we test is $1024$ to verify the ability of long-term gameplay, which is much greater than previous works (e.g., $64$ in GameNGen).
The results are shown in Table~\ref{tab:metrics}. When the prediction length is less than 128, we can see that PlayGen can achieve sufficient visual quality on metrics of LPIPS, PSNR, FID, FVD for \emph{Super Mario Bros} and PSNR, FID for \emph{Doom}. In particular, the PSNR of PlayGen can be higher than $30$ for \emph{Super Mario Bros}, showing great image quality \citep{dip}.
It is hard to distinguish between the generated frames and the ground-truth frames as shown in Fig.~\ref{fig:prediction_vis}. 

Besides, as the prediction length
increases, the visual quality metrics are getting worse, which is inevitable due to the accumulation of prediction errors. 
Noted that, because of the randomness of the environment, it is reasonable that generated frames and ground-truth frames differ significantly when the generated sequence is too long. In this case, using metrics such as LPIPS, PSNR, FID, and FVD to evaluate visual quality may yield poorer results. Thus, these metrics cannot fully reflect the visual quality when the prediction length is too long, which is the limitation of our evaluation method. Nevertheless, the difference in video quality metrics between a prediction length of 1024 and a prediction length of 128 is not substantial, which demonstrates that PlayGen can maintain adequate visual quality after long-term predictions.

\subsubsection{Accuracy of Game Interactive Mechanics}

As a playable game, it is important to accurately simulate interactive mechanics.
All previous related works \citep{genie,mariovgg,gameNgen,oasis} lack the evaluation regarding the simulation of game interactive mechanics. As mentioned in Sec.~\ref{sec:model_evaluation}, we propose two complementary action-aware metrics (i.e., ActAcc and ProbDiff) to assess the accuracy of the simulation of game interactive mechanics.
The VAM can only evaluate generated trajectories over 17 frames according to the experimental setup. It is not applicable for prediction length of $1$ and $16$.
Similar to the visual quality evaluation, we calculate two action-aware metrics on $600$ ground-truth trajectories.

By playing the game-generative model manually, we find that it has a great simulation of game interactive mechanics when $\text{AccAct} > 0.75$ and $\text{ProbDiff} < 0.1$ . Table~\ref{tab:metrics} shows the values of two action-aware metrics of PlayGen with different prediction lengths. As we can see, the AccAct is greater than $0.789$ and the ProbDiff is lower than $0.065$ with all prediction lengths on both games, which indicates that PlayGen can achieve precise simulation of game interactive mechanics. More importantly, the reduction in the accuracy of interactive mechanics is very small (less than 0.036 for ActAcc and 0.009 for ProbDiff) when we increase the prediction length from $32$ to $1024$. This confirms that PlayGen can maintain an accurate simulation of game interactive mechanics after long-term gameplay.

\subsubsection{Interaction Efficiency}

We finally test the interaction efficiency of PlayGen. We utilize the number of interaction frames per second (i.e., FPS) to reflect the interaction efficiency. Generally speaking, $20$ FPS can be considered as being able to achieve real-time interaction. As shown in Table~\ref{tab:denoise}, we test PlayGen with different denoise
sampling timesteps on NVIDIA RTX 2060 GPU. It can be observed that PlayGen can satisfy the requirement of real-time interaction (i.e., 20 FPS) by setting $4$ denoise sampling timesteps, while meeting the playability criteria of most metrics (all metrics are satisfied for \emph{Super Mario Bros}, only LPIPS and FVD are not strictly satisfied for \emph{Doom}).

\subsection{Analysis}

\begin{figure}[t]
    \centering
    \begin{minipage}[!t]{1\linewidth}
        \subfloat[A case with poor simulation of game interactive mechanics. The results of action-aware metrics are \textbf{ActAcc = 0.3 and ProbDiff = 0.411}. 
        ]{
        \includegraphics[width=0.99\linewidth]{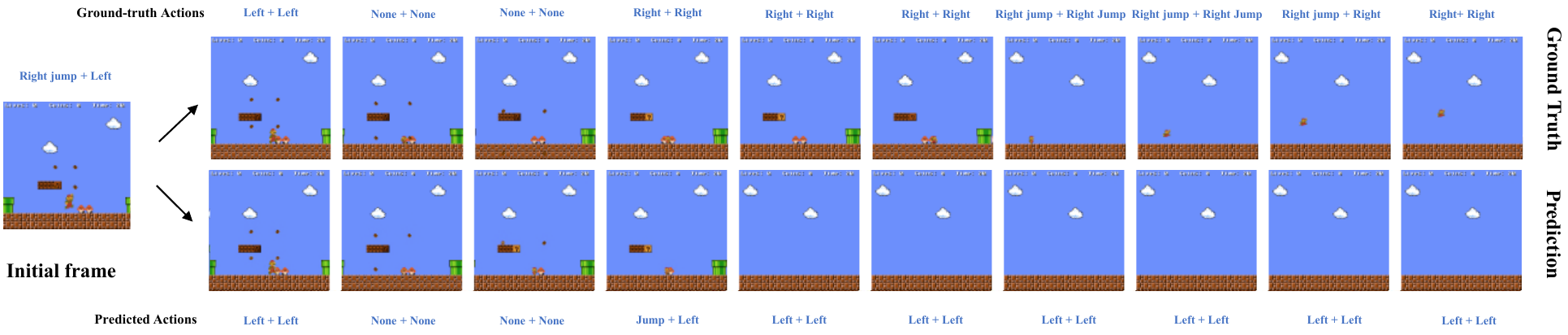}}
        \centering
    \end{minipage}
    \vspace{1.5em}
    
     \begin{minipage}[!t]{1\linewidth}
        \subfloat[A case with great simulation of game interactive mechanics. The results of action-aware metrics are \textbf{ActAcc = 1 and ProbDiff = 0}. 
        ]{ 
        \includegraphics[width=0.99\linewidth]{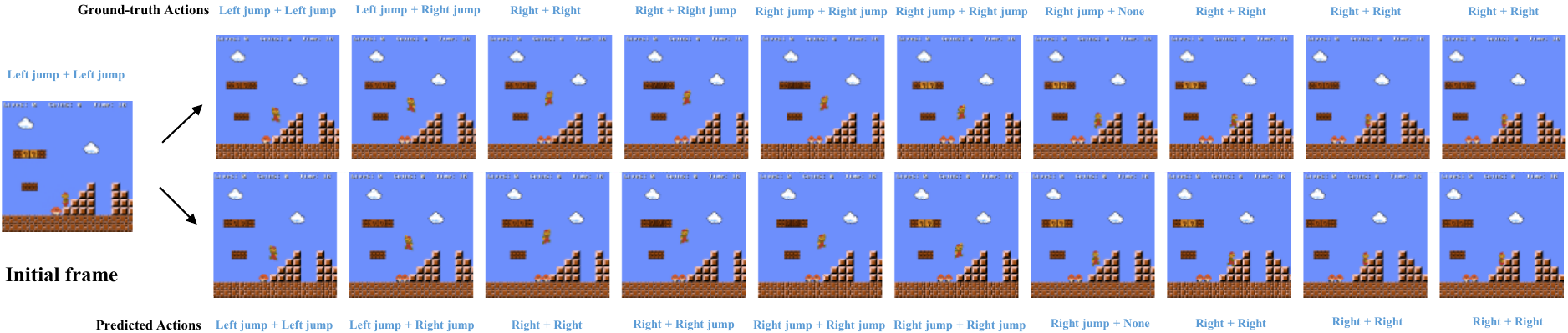}}
        \centering 
    \end{minipage}   

     \begin{minipage}[!t]{1\linewidth}
         \subfloat[A case with great simulation of game interactive mechanics. The results of action-aware metrics are \textbf{ActAcc = 0.6 and ProbDiff = 0.089}. 
         ]{ 
         \includegraphics[width=0.99\linewidth]{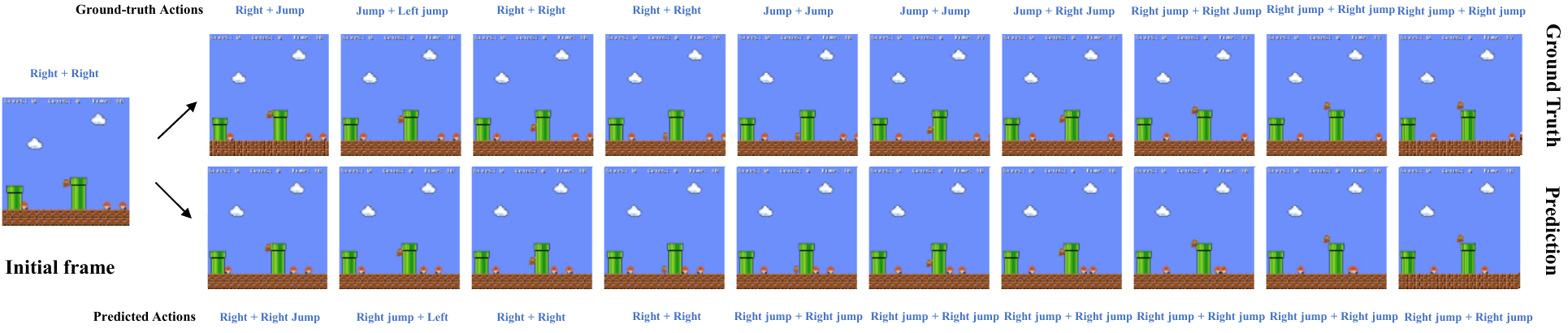}}
         \centering 
     \end{minipage}    
     \caption{\small \textbf{Cases demonstrating the rationality of action-aware metrics.} The values of ActAcc metric and ProDiff metric are highlighted in bold.}
     \label{fig:action_case}
  \vspace{-1em}
 \end{figure}

In this subsection, we analyze how PlayGen works. We first present specific cases in Fig.~\ref{fig:case_analysis} to demonstrate how we achieve playability step by step with different components of PlayGen. 
Then, we show the rationality of action-aware metrics by providing visualization results of the action-aware metrics in Fig.~\ref{fig:action_case}. 
Finally, we conduct experiments to analyze the impact of balanced data sampling.

\subsubsection{How PlayGen Progressively Achieves Playability}
\label{sec:case_analysis}

The most challenging aspect of ensuring ``playability'' is accurately simulating the game interactive mechanics. Hence, we show the progressive improvement of playability mainly on this part. 
In the following, we utilize specific cases on \emph{Super Mario Bros} to intuitively show how PlayGen becomes playable step by step based on three important components: diverse data collection, balanced data sampling and self-supervised long-tailed transition learning.

\begin{figure}[t]
    \centering
    \begin{minipage}[!t]{1\linewidth}
        \subfloat[The comparison of generated results before (top row) and after (bottom row) using \textbf{diverse data collection}. 
        ]{
        \includegraphics[width=0.99\linewidth]{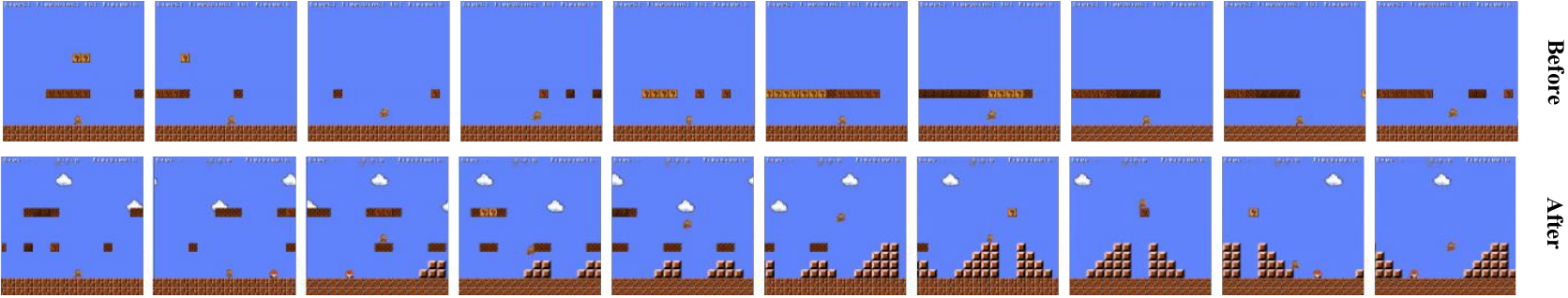}}
        \centering
    \end{minipage}
    \vspace{1.5em}
    
    \begin{minipage}[!t]{1\linewidth}
        \subfloat[The comparison of generated results before (top row) and after (bottom row) using \textbf{balanced data sampling}. 
        We use red boxes to highlight Mario to better illustrate its movement trajectory.
        ]{ 
        \includegraphics[width=0.99\linewidth]{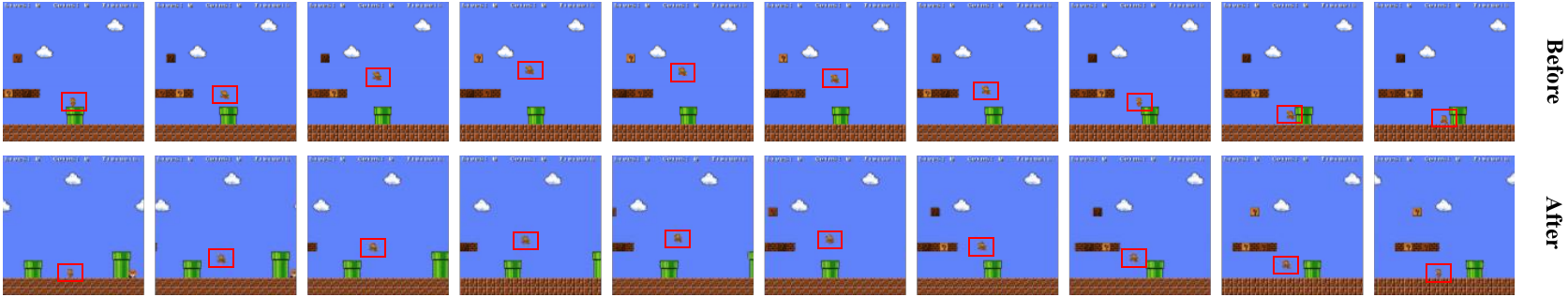}}
        \centering 
    \end{minipage}
    \vspace{1.5em}

    \begin{minipage}[!t]{1\linewidth}
        \subfloat[The comparison of generated results before (top row) and after (bottom row) using \textbf{self-supervised long-tailed transition learning}. 
        We use red boxes to highlight Mario and the mushroom in the frames before and after consuming the mushroom, for better illustration.
        ]{ 
        \includegraphics[width=0.99\linewidth]{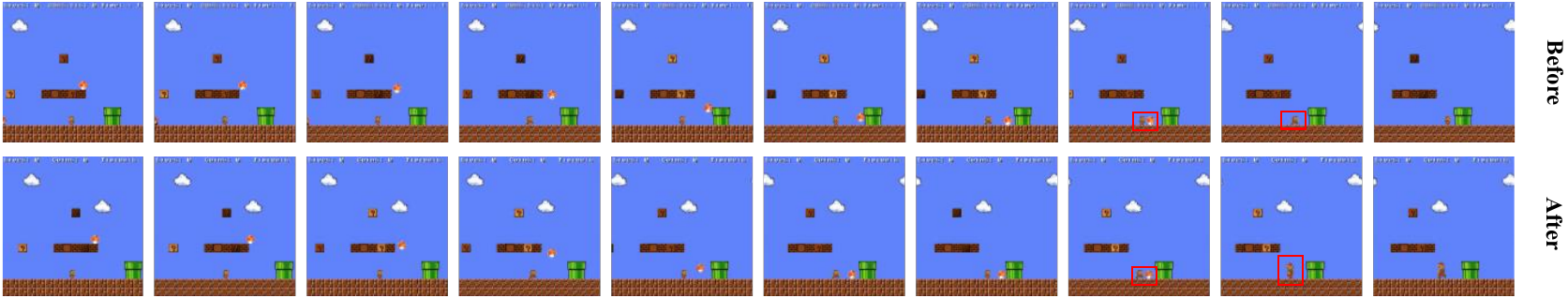}}
        \centering 
    \end{minipage}    
    \caption{
\textbf{\small Cases of how PlayGen incrementally improves playability.} The components of PlayGen that enhance playability in each case are highlighted in bold.
    }
    \label{fig:case_analysis}
\vspace{-1em}
\end{figure}

\paragraph{Diverse Data Collection.} We first utilize the diverse data collection method to improve the transition coverage. Fig. \ref{fig:case_analysis}(a) illustrates the comparison of gameplay results before and after using this method. It can be seen that without diverse data collection, the generated frames have incomplete backgrounds, missing many elements such as clouds, monsters, and pipes, and there is an unreasonable sequence of bricks, which is due to insufficient transition coverage.
After using this method, we are able to collect data covering almost all transitions in the game. Hence, we can observe that the backgrounds of generated frames are significantly richer and adhere to the game interactive mechanics. This demonstrates the importance of the diverse data collection method for improving playability.

\paragraph{Balanced Data Sampling.} 
With diverse data collection, we solved the transition coverage problem. However, it is still not playable.
As shown in the top row of frames in Fig. \ref{fig:case_analysis}(b), Mario performs a ``left jump'' action, but appears to be obstructed, resulting in a very short jump distance, which does not conform to the game interactive mechanics. This is because when we collect data, the agent will perform more ``move right'' and ``right jump'' actions to complete the game, which results in very little data going left compared to data going right, i.e., imbalanced data distribution. We address this problem with the balanced data sampling method. After balancing the data distribution, the generated game executes the action correctly as shown in the bottom row in Fig. \ref{fig:case_analysis}(b), making PlayGen more playable.


\paragraph{Self-Supervised Long-Tailed Transition Learning.} 
Now, we can simulate common game interactive mechanics, but still fail in some rare transitions. As shown in the top row of frames in Fig. \ref{fig:case_analysis}(c), Mario does not grow bigger after consuming a mushroom, which violates interactive mechanics. We attribute this to the fact that the proportion of mushroom-consuming transitions is particularly low, making it a long-tailed problem. We tackle this with the self-supervised long-tailed transition learning method. The bottom row in Fig. \ref{fig:case_analysis}(c) shows the results after using this. It can be seen that Mario grows bigger after consuming a mushroom, indicating we solve the long-tailed transition problem, which allows PlayGen to ultimately achieve a high level of playability.

 \begin{table}[t]
    \centering
    \renewcommand{\arraystretch}{1.2}
    \setlength{\tabcolsep}{6pt}
    \caption{A comparative analysis of metrics evaluating the visual quality and accuracy of game interactive mechanics in game-generative models trained with and without \textbf{balanced data sampling}. The best results for each metric are highlighted in bold.}
    \begin{tabular}{c|c c|c c}
        \toprule
        \multirow{2}{*}{Metric} & \multicolumn{2}{c|}{\emph{Super Mario Bros}} & \multicolumn{2}{c}{\emph{Doom}} \\
        \cline{2-5}
        & w/o & w/ & w/o & w/ \\
        \hline
        LPIPS$\downarrow$ & 0.081 & \textbf{0.062} &  0.457 & \textbf{0.285} \\
        PSNR$\uparrow$ & 23.89 & \textbf{26.18} & 17.45 & \textbf{20.41} \\
        FID$\downarrow$ & 10.01 & \textbf{6.12} & 107.53 & \textbf{58.75} \\
        FVD$\downarrow$ & 195.59 & \textbf{105.44} & 1470.17 & \textbf{622.51} \\
        ActAcc$\uparrow$ & 0.793 & \textbf{0.801} & 0.842 & \textbf{0.858} \\
        ProbDiff$\downarrow$ & 0.062 & \textbf{0.059} & 0.024 & \textbf{0.019} \\
        \bottomrule
    \end{tabular}
    \label{tab:balanced_sample}
\end{table}

\subsubsection{Rationality of Action-Aware Metrics}

PlayGen uses two action-aware metrics ActAcc and ProbDiff to evaluate the game interactive mechanics. Below, we show the rationality of these metrics with three cases. 

Fig. \ref{fig:action_case}(a) is a case that presents a poor simulation of game interactive mechanics. Specifically, in the ground-truth frames, Mario shrinks after encountering a Goomba and then continues to jump forward. In the generated frames, however, the game-generative model produces an error, resulting in Mario’s absence from the scene. The ActAcc metric value ($0.3$) is very low, while the ProbDiff metric value ($0.411$) is very high, which demonstrates that our action-aware metrics successfully identified this as a bad case. 

The second case presents a great simulation of game interactive mechanics as shown in Fig. \ref{fig:action_case}(b). The generated frames are identical to ground-truth frames, reflecting an unbiased simulation of the interactive mechanics, which aligns with the results of our action-aware metrics (ActAcc = 1 and ProbDiff = 0). 


The last case also presents an excellent simulation of game interactive mechanics as shown in Fig. \ref{fig:action_case}(c).  We observe that when Mario is airborne, the actions do not influence subsequent frames. Consequently, the action probability distribution predicted by VAM becomes more uniform, resulting in a low ActAcc metric value ($0.6$), which can mislead the evaluation of interactive mechanics. In contrast, since ProbDiff measures the differences in probabilities between actions, it remains unaffected by this scenario and stays at a low value ($0.089$). This indicates that the case adheres well to the game’s interactive mechanics. Therefore, ProbDiff serves as a crucial metric to complement ActAcc in the assessment of game interactive mechanics.

Overall, these cases illustrate that the two action-aware metrics can effectively complement each other in evaluating the accuracy of interactive mechanics, in concordance with human visual assessments.

\subsubsection{The Impact of Balanced Data Sampling}
As demonstrated in  Sec.~\ref{sec:case_analysis}, balanced data sampling is a crucial component that significantly enhances the playability of PlayGen. Here, we present the quantitative impact of balanced data sampling.
Table \ref{tab:balanced_sample} compares the performance of game-generative models trained using the original dataset (236M frames for \emph{Super Mario Bros} and 900M frames for \emph{Doom}) with those trained on the balanced dataset (50M frames for \emph{Super Mario Bros} and 200M frames for \emph{Doom}). 
The results indicate that balanced data sampling leads to improvements ranging from 1\% to 46.1\% across various metrics for the \emph{Super Mario Bros} game, and from 1.9\% to 57.7\% for the \emph{Doom} game.

%% file: section/5_conclusion.tex
\section{Discussion} 

\paragraph{Summary}

%

In this work, we introduce PlayGen, a novel approach designed to tackle the playability challenges in game generation. Through extensive validation on both the 2D \emph{Super Mario Bros} and the 3D \emph{Doom}, PlayGen demonstrates its capability to provide real-time interaction at 20 FPS, maintain high visual quality with PSNR values of 33.81 for \emph{Super Mario Bros} and 23.81 for \emph{Doom}, and accurately simulate interactive mechanics. These achievements are consistently sustained across over 1000 frames of gameplay on an NVIDIA RTX 2060 GPU.
To foster further research and development within the community, we have made our code, trained models, and an interactive demo fully accessible and open-source.

\paragraph{Limitations}
The RNN-like diffusion model theoretically has infinite memory capacity, but practical limitations still exist. 
 The RNN architecture ensures that memory is retained, preventing rendering crashes. However, excessively long memories can become inaccurate, leading to hallucinations.
For example, in the \emph{Doom} game, if there are two very similar paths leading to two different rooms, choosing the first path might result in occasionally ending up in the second room. 
This issue arises because distinguishing between similar historical information is challenging, necessitating precise memory to achieve correct outcomes. While the RNN structure guarantees memory length, it falls short in providing the precision required for long-term memory accuracy.



\paragraph{Future Work}
1) Conducting more ablation studies on playability. We provide an intuitive analysis of the process of achieving playability but lacking quantitative results. 
We plan to conduct additional quantitative ablation studies to better understand this process, focusing on the impact of data coverage, long-tailed transition learning, and model architecture.
2) Exploring more complex game scenarios. Although \emph{Super Mario Bros} and \emph{Doom} are representative games in 2D and 3D game scenarios respectively, their visual quality are relatively  low and interactive mechanics are simple. It would be intriguing  to leverage PlayGen to generate games with more complex interactive mechanics and higher resolution, akin to AAA games. 3) Supporting prompt customization. Currently, PlayGen is capable of generating games frame by frame through actions. We aim to further develop PlayGen to generate games controllable by prompts, similar to current generative models. To achieve this, we plan to expand the model structure and create a prompt dataset.

%% file: section/appendix.tex
\appendix

\section{More Implementation Details}

\subsection{Model Hyperparameter Setting}
We provide an overview of the hyperparameters of VAE and LDM models in Tab.~\ref{tab:hyperparameters}.

\begin{table}[h]
    \caption{Hyperparameters for VAE and LDM models.}
    \centering
    \begin{minipage}{0.45\linewidth}
        \renewcommand\arraystretch{1}
        \centering
        \subfloat[VAE]{
        \resizebox{1\textwidth}{!}{%
        \begin{tabular}{lc}
            \toprule
            Item                   & Value         \\ \hline
            Batch Size             & 32            \\
            Parameters             & 55M           \\
            Channels               & 128           \\
            Channel Multiplier     & {[}1, 2, 4{]} \\
            Latent Shape           & 4$\times$32$\times$32       \\
            ResNet Block Number    & 2             \\
            KL Loss Weight         & 1e-6          \\
            Perceptual Loss Weight & 1             \\
            Learning Rate          & 4.5e-6        \\
            \bottomrule
            \end{tabular}}}
    \end{minipage}
    \hspace{1mm}
    \begin{minipage}{0.51\linewidth}
        \centering
        \subfloat[LDM]{
        \resizebox{1\textwidth}{!}{%
        \begin{tabular}{lc}
            \toprule
            Item                        & Value    \\ \hline
            Batch Size                  & 8        \\
            Parameters                  & 131M      \\
            $z$-shape                   & 32$\times$32$\times$32 \\
            Diffusion Steps             & 1000     \\
            Noise Schedule              & sigmoid  \\
            Sequence Length             & 36       \\
            DiT Depth                   & 12       \\
            DiT Hidden Size             & 384      \\
            DiT Patch Size              & 2        \\
            DiT Num Heads               & 6        \\
            Time Embedding Dimension   & 192      \\
            Action Embedding Dimension & 192      \\
            Learning Rate               & 1e-4    \\
            \bottomrule
        \end{tabular}}}
    \end{minipage}
\label{tab:hyperparameters}
\end{table}


\subsection{Data Collection}

\paragraph{Super Mario Bros.}
When generating data for the \emph{Super Mario Bros} game, we collect 200 timesteps per sample. After exceeding 200 timesteps, regardless of whether the agent dies or wins, we truncate and save the sample. At the beginning of each sample, the agent is spawned at a random position on the map with a random status (small, large, or fire) to enhance map coverage. If the agent dies or wins during the episode, it respawns at the initial point of the map in the small status, in accordance with the rules of game.

\paragraph{Doom.}
We also collect 200 timesteps per sample in \emph{Doom}. Similar to \emph{Super Mario Bros}, we place the agent at a random position on the map at the beginning of each sample. Additionally, at the start of each episode (from the beginning of the game to death/victory), we randomly select an action as the preferred action for the current episode. This preferred action will have a higher probability of being chosen during the episode.

\section{More Visualization Results}
\begin{figure}[t]
    \centering
    \begin{minipage}[!t]{1\linewidth}
        \subfloat[\emph{Super Mario Bros}
        ]{
        \includegraphics[width=0.99\linewidth]{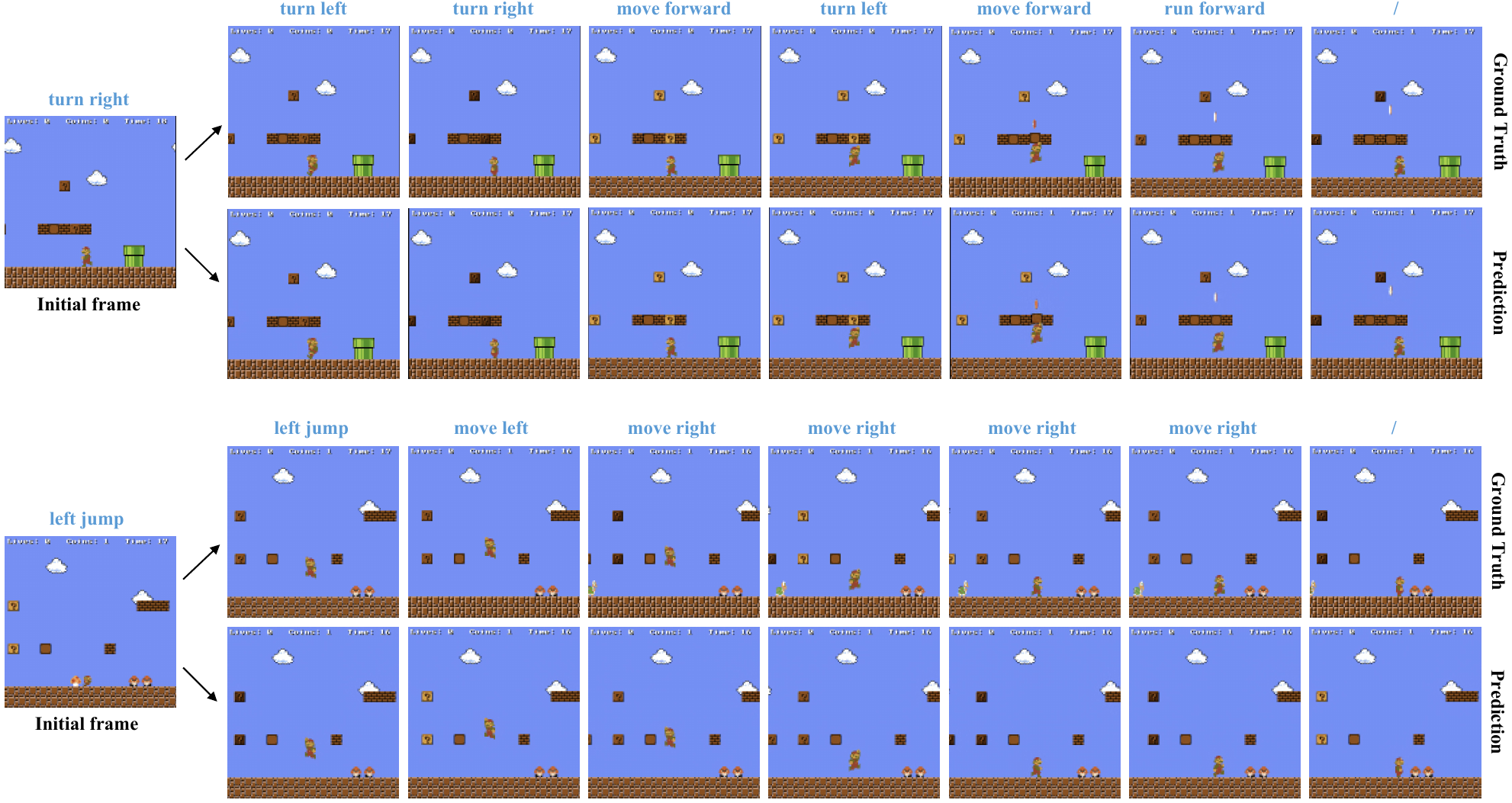}}
        \centering
    \end{minipage}
    \vspace{1.5em}

    \begin{minipage}[!t]{1\linewidth}
        \subfloat[\emph{Doom}
        ]{ 
        \includegraphics[width=0.99\linewidth]{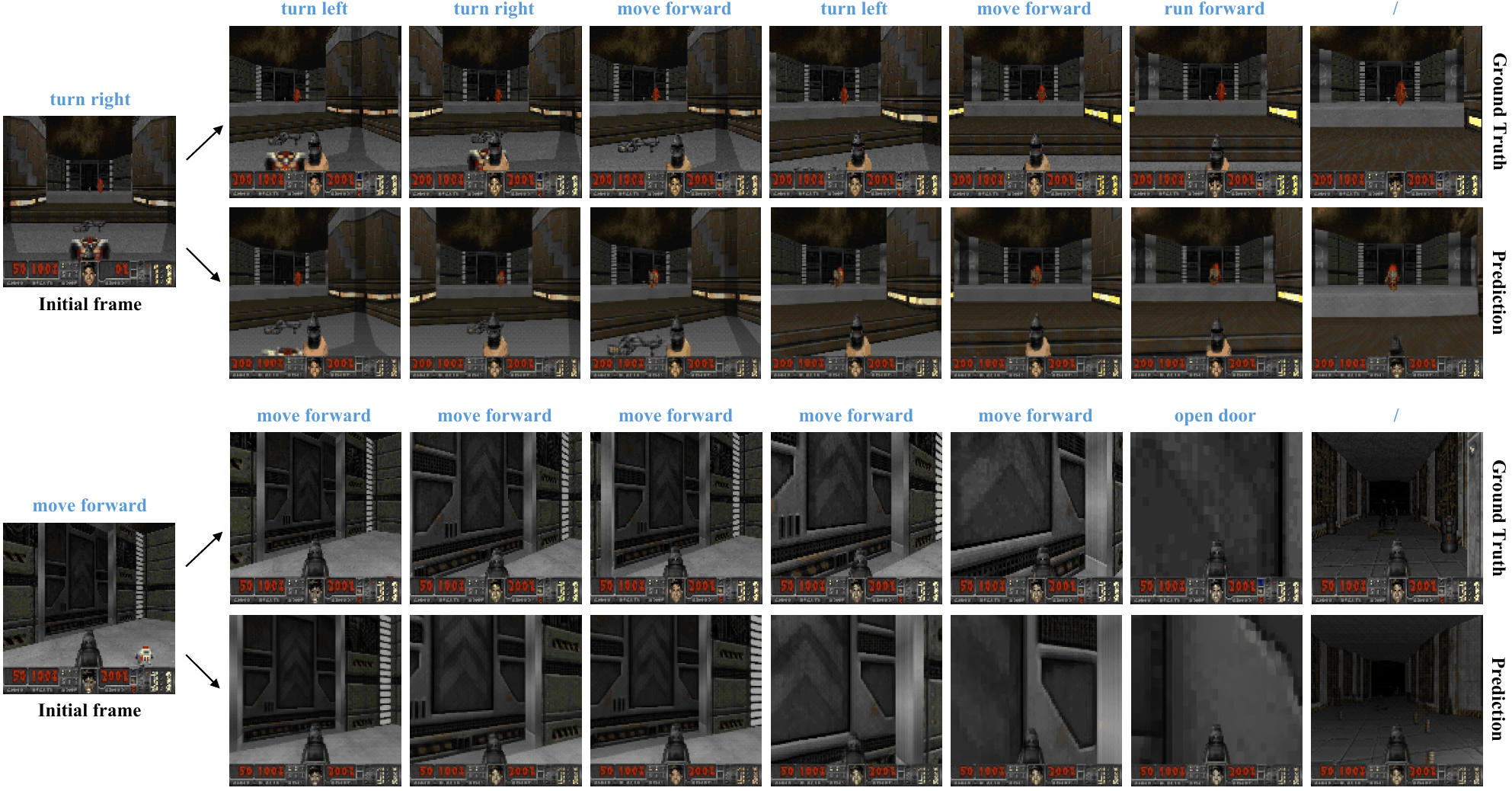}}
        \centering 
    \end{minipage}    
    \caption{
    More visualization results that predicted by PlayGen within \emph{Super Mario Bros} and \emph{Doom}.
    }
    \label{fig:supp_vis}
\end{figure}

We provide more visualization results in \emph{Super Mario Bros} and \emph{Doom} in Fig.~\ref{fig:supp_vis}. PlayGen generate the subsequent frames according to the initial frame and given actions.

